\documentclass{vldb}
\usepackage{graphicx}
\usepackage{semantic}

\usepackage{amsthm}
\usepackage{amssymb}
\usepackage{multirow}
\usepackage{url}
\usepackage{hyperref}
\graphicspath{ {images/} }
\usepackage{url}
\usepackage{flushend}

\begin{document}
\sloppy
\title{Logical Inferences with Contexts of RDF Triples}
\numberofauthors{2} 
\author{
\alignauthor
Vinh Nguyen\\
       \affaddr{Kno.e.sis Center}\\
       \affaddr{Wright State University}\\
       \affaddr{Dayton, Ohio, USA}\\
       \email{vinh@knoesis.org}
\alignauthor Amit Sheth\\
        \affaddr{Kno.e.sis Center}\\
       \affaddr{Wright State University}\\
       \affaddr{Dayton, Ohio, USA}\\
       \email{amit@knoesis.org}
}
\maketitle
\pagestyle{empty}

\begin{abstract}

Logical inference, an integral feature of the Semantic Web, is the process of deriving new triples by applying entailment rules on knowledge bases. The entailment rules are determined by the model-theoretic semantics. 
Incorporating context of an RDF triple (e.g., provenance, time, and location) into the inferencing process requires the formal semantics to be capable of describing the context of RDF triples also in the form of triples, or in other words, \textit{RDF contextual triples about triples}. The formal semantics should also provide the rules that could entail new contextual triples about triples. 


In this paper, we propose the first inferencing mechanism that allows context of RDF triples, represented in the form of RDF triples about triples, to be the first-class citizens in the model-theoretic semantics and in the logical rules. Our inference mechanism is well-formalized with all new concepts being captured in the model-theoretic semantics. This formal semantics also allows us to derive a new set of entailment rules that could entail new contextual triples about triples.

To demonstrate the feasibility and the scalability of the proposed mechanism, we implement a new tool in which we transform the existing knowledge bases to our representation of RDF triples about triples and provide the option for this tool to compute the inferred triples for the proposed rules. We evaluate the computation of the proposed rules on a large scale using various real-world knowledge bases such as Bio2RDF NCBI Genes and  DBpedia. The results show that the computation of the inferred triples can be highly scalable. On average, one billion inferred triples adds 5-6 minutes to the overall transformation process. NCBI Genes, with 20 billion triples in total, took only 232 minutes for the transformation of 12 billion triples and added 42 minutes for inferring 8 billion triples to the overall process.




\end{abstract}

\section{Introduction}
\label{introduction}
Semantic Web technologies such as RDF and OWL are emerging as standard languages for machine-understandable knowledge representation and reasoning. A Semantic Web knowledge base, as a set of RDF triples, can be created using different methods. Existing data from a structured form (e.g., relational databases, text files, XML documents, or HTML pages) can be transformed into RDF with ontologies describing the database schema (e.g., Bio2RDF \cite{belleau2008bio2rdf} and PubChem \cite{pubchem-fu}). A knowledge base can also be created by extracting the triples in the form of (subject, predicate, object) from unstructured data using natural language processing algorithms (e.g., Google Knowledge Vault \cite{dong2014knowledge}, Yago2S \cite{hoffart2013yago2}, and  DBpedia \cite{lehmann2015dbpedia}). 
In either method, each RDF triple in the resulting knowledge bases can be associated and enriched with different types of contextual information, such as the time duration in which the triple holds true, and the provenance specifying the source of the triple. 

If a knowledge base is created and maintained by an organization, the knowledge integrity can be validated within that organization. 
Nowadays, it is commonplace for a knowledge base to be created and shared by anyone on the Web,
 or in the Linked Open Data. Therefore, we believe that the context of every fact or assertion should be provided for consumers to validate and assess the reliability of the knowledge before using it. The contextual information of a triple may provide the time interval or the time instant when the triple holds true so that the consumers can validate it. It may also provide the Web page or the article where the triple was extracted from, or any provenance information that allows for tracking the origin of the triples so that the consumers can assess the reliability of the triples. Since a fact would not hold true in every context, the context in which a proposition holds true needs to be presented in the knowledge bases so that the proposition can be validated and reused. 
 
 Popular knowledge bases currently represent the contextual information of their triples in the RDF quad form. For example,  DBpedia \cite{dbpedia}, CTD \cite{ctd}, and GO Annotations \cite{goa}, NCBI Genes \cite{ncbigene}, and PharmGKB \cite{pharmgkb} have the provenance of every RDF triple represented in the quad form. Since the fourth element of the RDF quad is not formalized as first-class citizen of the current model-theoretic semantics \cite{hayesrdf}, it cannot be represented in the entailment rules.
Therefore, to support the inferences involving contextual information about RDF triples, we believe that the current RDF and OWL semantics
should be expanded to allow for the accomodation of contexts of triples as first-class citizens in the model-theoretic semantics and entailment rules.


\subsection{Motivating Example}
 
Given a statement ``Barack Obama is married to Michelle Obama'' ($T_1$), and a sub-property relationship between \textit{isMarriedTo} and \textit{isSpouseOf} ($T_2$), applying the rule rdfs7 \cite{hayes2004rdf} for rdfs:subPropertyOf on the two triples will entail the new triple ``Barack Obama is a spouse of Michelle Obama'' ($T_3$).

\begin{tabular}{l@{\hspace{1em}}l @{\hspace{1em}}l @{\hspace{1em}}l}  \noalign{\smallskip}
$T_1$: & BarackObama & isMarriedTo & MichelleObama . \\
$T_2$: & isMarriedTo & subPropertyOf & isSpouseOf .\\
$T_3$: & BarackObama & isSpouseOf & MichelleObama . \\\noalign{\smallskip}
\end{tabular}

These triples would provide enough knowledge for answering simple questions. For example, who is Barack Obama married to? Or who is spouse of Barack Obama?

However, these triples do not provide sufficient knowledge to give answers to more complex questions. For example, when and where did Barack Obama marry Michelle Obama? Was he married to Michelle Obama before 2008? Was he a spouse of Michelle Obama in Harvard, Massachusetts?

To answer these questions, contextual information such as time and location need to be represented, as in ``Barack Obama is married to Michelle Obama in Chicago in 1992''. Given this contextual statement and the knowledge that Chicago is a city in Illinois, and Illinois is a state in the USA, we as humans can quickly infer many contextual statements which could answer the above questions. Furthermore, we can also infer statements that could be much different from the original statement and from each other. For example, the statements $S_1$ and $S_5$ in Table \ref{tab:derived_statements} only share the subject and the object while their relationships and contextual information are totally different, as shown in Table \ref{tab:derived_statements} with the inferred information in bold font. 

How does a machine become intelligent enough to infer contextual statements as humans do? We address this question by (1) developing an inferencing mechanism that could entail new contextual statements and (2) demonstrating an inferencing process that derives the human-inferred contextual statements from the original statement.

In order for machines to infer contextual statements, we believe that two requirements need to be fulfilled. First, these contextual statements must be represented in machine-understandable form, with explicitly defined semantics for the relationship between the triple and its contextual information. Second, we need an entailment mechanism that takes into account the semantics of the contextual triples about triples so that it can entail the new ones.

To the best of our knowledge, there is no RDF/RDFS inferencing rule that involves RDF contextual triples about triples. 
Our paper addresses this missing capability.

\begin{table}[t]
  \centering
\resizebox{0.48\textwidth}{!}{%
  \begin{tabular}{l} \hline\noalign{\smallskip}
    \textbf{Deriving statements} \\ \hline \noalign{\smallskip}
	$S_1$. BarackObama is married to  MichelleObama \textbf{in Illinois}.\\
	$S_2$. BarackObama is married to  MichelleObama \textbf{in USA}. \\ \hline\noalign{\smallskip}
  	$S_3$. \textbf{BarackObama became spouse of MichelleObama} in Chicago. \\ \hline \noalign{\smallskip}
    	$S_4$. \textbf{BarackObama became spouse of MichelleObama in Illinois}. \\
	$S_5$. \textbf{BarackObama became spouse of MichelleObama in USA}. \\ \hline\noalign{\smallskip}
    
  \end{tabular}}
  \caption{Statements inferred from ``BarackObama is married to MichelleObama in Chicago''}
  \label{tab:derived_statements}
\end{table}

\subsection{Approach}
The contextual statement ``Barack Obama is married to Michelle Obama in Chicago'' is used as an illustrative example throughout the paper. We represent this contextual statement and the background knowledge of Chicago in the form of triples as follows:

\begin{tabular}{l@{\hspace{1em}}l @{\hspace{1em}}l @{\hspace{1em}}l}  \noalign{\smallskip}
$T_1$: & BarackObama & isMarriedTo & MichelleObama .\\
$T_4$: & $T_1$ & happenedIn & Chicago .\\
$T_5$: & Chicago & partOf & Illinois .\\
$T_6$: & Illinois & partOf & USA . \\\noalign{\smallskip}
\end{tabular}

We call $T_1$ a \textit{primary triple} and $T_4$ a \textit{meta triple} about location.
The primary triple $T_1$ is a regular RDF triple, while the meta triple $T_4$ describes the context in which the primary triple holds true.

Considering the original statement represented in $T_1$ and $T_4$, there is no relationship between the triple ``BarackObama isMarriedTo MichelleObama'' and its identifier $T_1$. Therefore, there is no relationship between the primary triple $T_1$ and the meta triple $T_4$.

To bridge this gap and fulfill the requirement of representing contextual statements in machine-understandable form as discussed earlier, several approaches such as named graph \cite{carroll2005named}, RDF reification \cite{hayes2004rdf}, and singleton property \cite{Nguyen:2014:DLR:2566486.2567973} can be used for representing the relationship between a triple and its identifier. Furthermore, the semantics of the relationship between a triple and its identifier must be captured and expressed as a first-class citizen in the formal semantics. It allows the semantics of the contextual statements to be expressed in the formal model as well as in the logical rules.

Among the existing approaches, the singleton property representation comes with a formal semantics formalizing the relationship between the singleton property and the triple it represents. Meanwhile, although the named graph could be used as a triple identifier, it was not intended to be used for that purpose. Instead, the named graph is mainly used for representing a set of triples. The RDF reification does not have a formal semantics for capturing the relationship between a statement instance and the triple it represents. As a result, we choose the singleton property (SP) representation and use its semantics in our entailment mechanism. Correspondingly, all the knowledge bases available in the form of RDF reification or named graph need to be transformed into the singleton property representation for inference purposes.





This paper has two contributions:

\begin{itemize}
\item We developed a new entailment mechanism that allows contextual statements, represented in the form of RDF contextual triples about triples, to be the first-class citizens in model-theoretic semantics as well as in the logical rules. The proposed mechanism is well-formalized with all new concepts for RDF triples about triples (Section \ref{concepts}) being captured by a model-theoretic semantics (Section \ref{formal}). The formal semantics also allows us to derive a new set of rules that could entail new contextual statements (Section \ref{entailment-rules}). We demonstrate how the human-inferred contextual statements in Table \ref{tab:derived_statements} can be entailed using the proposed rules (Section \ref{reasoning-types}).
\item We demonstrated that the proposed entailment mechanism is scalable. We developed a new tool, called rdf-contextualizer, to transform existing knowledge bases from the named graph form to the singleton property representation. We then provide the option to compute all inferred triples using the proposed contextual entailment rules in this tool (Section \ref{implementation}). We evaluated the computational performance in very large real-world knowledge bases with billions of triples such as  DBpedia and NCBI Genes (Section \ref{evaluation}).
\end{itemize}

This paper presents the foundational capability. While the location context and simple question answering are used as the illustrative examples, discussions of a broad class of Semantic Web applications such as querying SPARQL with backward-chaining reasoning, streaming reasoning, temporal reasoning, tracking context of inferred triples, and question answering based on extracted knowledge and logical inferences, are beyond the scope of the paper.

The remaining sections are as follows. 
We present the related work in Section \ref{related}. We discuss the future work in Section \ref{discuss} and conclude with Section \ref{conclude}.


\section{Conceptual Model}
\label{concepts}

\subsection{Preliminaries}
\label{background}


Here we recall the singleton property concept with its syntax and semantics from \cite{Nguyen:2014:DLR:2566486.2567973}.

A \textit{singleton property} is a specific property instance that represents a unique relationship under a specific context. 

For the example at hand, the singleton property \textit{isMarriedTo\#1} uniquely represents the \textit{isMarriedTo} relationship between BarackObama and MichelleObama. 
This singleton property can be asserted with contextual information ``in Chicago'' about the relationship as follows:

\resizebox{0.47\textwidth}{!}{%
\begin{tabular}{l@{\hspace{1em}}l @{\hspace{1em}}l @{\hspace{1em}}l}  \noalign{\smallskip}
$SP_1$: & BarackObama & isMarriedTo\#1 & MichelleObama .\\
$SP_2$: & isMarriedTo\#1 & singletonPropertyOf & isMarriedTo .\\
$SP_3$: & isMarriedTo\#1 & happenedIn & Chicago .\\ \noalign{\smallskip}
\end{tabular}}

\textbf{Formal semantics}. 
Here we recall the mapping function $I_{EXT}$ from the current model-theoretic semantics \cite{hayesrdf}.

A property mapping function $I_{EXT}$ is a binary relation that maps one property to a set of pairs of resources.

Formally, let IP be the set of properties, IR be the set of resources. Then
$I_{EXT}$: IP $\rightarrow$ $2^{IR \times IR}$.

A singleton mapping function $I_{S\_{EXT}}$ is a binary relation that maps one singleton property to one pair of resources.

Formally, let IR be the set of resources, IPs $\subseteq$ IP be the set of singleton properties. Then
$I_{S\_{EXT}}$: IPs $\rightarrow$ IR $\times$ IR.

For example, $I_{S\_{EXT}}$(isMarriedTo\#1) = $\langle$BarackObama, MichelleObama$\rangle$.
As the singleton property \textit{isMarriedTo\#1} is also a property, its property extension is a singleton set, which has only one element. \\
$I_{EXT}$(isMarriedTo\#1) = {$\langle$BarackObama, MichelleObama$\rangle$}.

The syntax and the semantics of singleton properties described here are sufficient to allow us to attach contextual information for any RDF individual triple.

Next, we describe how singleton properties can be utilized for developing our inference scheme to infer new triples.

\subsection{Property Types}

A \textit{generic property} asserts the relationship between the subject and the object without providing additional contextual information about the relationship. This property groups all singleton properties sharing the same characteristics across contexts, and it is connected to a singleton property via the property \verb|singletonPropertyOf|.

\resizebox{0.47\textwidth}{!}{%
\begin{tabular}{l@{\hspace{1em}}l @{\hspace{1em}}l @{\hspace{1em}}l}  \noalign{\smallskip}
$SP_2$: & isMarriedTo\#1 & singletonPropertyOf & isMarriedTo .\\ \noalign{\smallskip}
\end{tabular}}

In this example, \verb|isMarriedTo| is a generic property.

Here we propose to add the new class \verb|GenericProperty| to represent the set of generic properties, in addition to the class \verb|SingletonProperty| from \cite{Nguyen:2014:DLR:2566486.2567973}. Any property that is not defined as a singleton property can become generic property. Intuitively, a generic property is mapped to a set of pairs while the singleton property is mapped to a single pair. If a singleton property also plays the role of a generic property, it will occur as the predicate in multiple triples. This contradicts to the definition of the singleton property. Therefore, a singleton property should never be defined as a generic property. In other words, the set of generic properties and the set of singleton properties are disjoint. The two classes \verb|SingletonProperty| and \verb|GenericProperty| do not share common instances.

Every generic property is an RDF property, but not every RDF property is a generic property as, for example, it could be a singleton. That makes \verb|GenericProperty| a sub-class of \verb|Property|.

\resizebox{0.42\textwidth}{!}{%
\begin{tabular}{l@{\hspace{1em}}l @{\hspace{1em}}l @{\hspace{1em}}l}  \noalign{\smallskip}
 rdf:GenericProperty & rdfs:subClassOf & rdf:Property & .\\
 rdf:SingletonProperty & rdfs:subClassOf & rdf:Property & .\\\noalign{\smallskip}
\end{tabular}}

Every generic property is an instance of the \verb|Property| class. However, one \verb|Property| instance may not belong to any of the classes \verb|SingletonProperty| or \verb|GenericProperty|. We call this a {\it regular property}. In the example at hand, \verb|partOf| is a regular property.

Although a regular property in some cases may share the same property extension with a generic property, it is necessary to make the clear distinction between them. 

\textbf{Distinguishing property types}. Here we distinguish three types of properties: singleton, generic, and regular. We also call them context-associated, context-dissociated, and context-agnostic property, respectively.
A singleton property such as \verb|isMarriedTo#1| is called context-associated because it can be asserted with contextual information. A generic property, or context-dissociated property such as \verb|isMarriedTo|, does not have contextual information attached. Finally, a regular property such as \verb|happenedIn|, instance of \verb|Property| class, is called a context-agnostic property because it is not yet committed to be associated or dissociated with contextual information.



\subsection{Triple Types}
\label{triple}


Similar to the distinction of the property types, here we can also distinguish the type of a triple based on the type of its property. The three types of properties form three types of triples: singleton, generic, and regular triple. 
A singleton triple is the only triple that has its singleton property occurring as a predicate.
A generic triple is a triple that has its predicate asserted as a generic property.
A regular triple is a triple that does not have its predicate asserted as a generic or singleton property.

\resizebox{0.47\textwidth}{!}{%
\begin{tabular}{l@{\hspace{1em}}l @{\hspace{1em}}l @{\hspace{1em}}l}  \noalign{\smallskip}
$Singleton$: & BarackObama & isMarriedTo\#1 & MichelleObama .\\
$Generic$: & BarackObama & isMarriedTo & MichelleObama .\\
$Regular$: & Chicago & partOf & Illinois .\\ \noalign{\smallskip}
\end{tabular}}

\textbf{Distinguishing triple types based on context.} The three types of properties form three types of triples: context-associated, context-dissociated, and context-agnostic triple. 
A context-associated triple is a singleton triple that may have contextual information associated with it via its singleton property.
A context-dissociated triple is a generic triple that does not have any contextual information associated with it.
A context-agnostic triple is a regular triple that is not committed to be associated or dissociated with contextual information.

\subsection{Contextual triple instantiation} 

Back to the motivating example, we have the original statement ``Barack Obama married to Michelle Obama in Chicago'' represented as follows:

\begin{tabular}{l@{\hspace{1em}}l @{\hspace{1em}}l @{\hspace{1em}}l}  \noalign{\smallskip}
$T_1$: & BarackObama & isMarriedTo & MichelleObama .\\
$T_4$: & $T_1$ & happenedIn & Chicago .\\ \noalign{\smallskip}
\end{tabular}

We utilize the singleton property graph pattern to bridge the gap between the primary triple $T_1$ and its identifier as explained in Section \ref{background}. 

\resizebox{0.47\textwidth}{!}{%
\begin{tabular}{l@{\hspace{1em}}l @{\hspace{1em}}l @{\hspace{1em}}l}  \noalign{\smallskip}
$SP_1$: & BarackObama & isMarriedTo\#1 & MichelleObama .\\
$SP_2$: & isMarriedTo\#1 & singletonPropertyOf & isMarriedTo .\\
$SP_3$: & isMarriedTo\#1 & happenedIn & Chicago .\\ \noalign{\smallskip}
\end{tabular}}

This singleton property graph pattern represents the primary triple $T_1$ and its meta triple $T_4$ by creating a singleton property \verb|isMarriedTo#1| and using it as triple identifier for asserting meta triples.

From triple $SP_2$, \verb|isMarriedTo#1| is the singleton property and \verb|isMarriedTo| is a generic property.
According to the triple type classification in Section \ref{triple}, $T_1$ is a generic triple and $SP_1$ is a singleton triple. 
$SP_1$ can be considered as one contextual triple instance of $T_1$. We generalize this relationship between the two triples as follows.

\textbf{Definition}. Let $sp$ be the singleton property of property $p$ and ($s$, $sp$, $o$) be the singleton triple, ($s$, $sp$, $o$) is the \textit{contextual triple instance} of the generic triple ($s$, $p$, $o$).

How a generic triple ($s$, $p$, $o$) can be derived from the singleton graph pattern will be formalized in the RDF formal semantics described in Section \ref{rdf}.

\section{Model-Theoretic Semantics}
\label{formal}

\subsection{ Mapping Functions}
\label{contextual_mapping}

In the formal semantics, we use the concept of function to assign the set of URIs and literals into the set of resources in a model interpretation, and assign each property a set of pairs (subject, object). 









Next, we define new mapping functions to be used in the interpretations described later in this section.
We reuse the set of singleton properties {\it IPs}, the set of properties {\it IP}, the mapping function $I_{EXT}$, and the singleton mapping function $I_{S\_{EXT}}$ mentioned in Section \ref{background}.

\textbf{Definition}. A generic property instance function $I_G$ is a binary relation that maps a generic property to a set of its singleton properties.

Formally, let {\it IPg} $\subseteq$ {\it IP} be the set of generic properties. We define the function 

$I_G:$ {\it IPg} $\rightarrow 2^{\it IPs}$ such that 

$I_G (p_g)= \{p_s \, | \, \langle p_s, p_g \rangle \in I_{EXT}($rdf:singletonPropertyOf$)\}$.

For example, \verb|isMarriedTo| is a generic property and it has two singleton properties as follows:

\resizebox{0.47\textwidth}{!}{%
\begin{tabular}{l @{\hspace{1em}}l @{\hspace{1em}}l}  \noalign{\smallskip}
isMarriedTo\#1 & rdf:singletonPropertyOf & isMarriedTo .\\
isMarriedTo\#2 & rdf:singletonPropertyOf & isMarriedTo .\\ \noalign{\smallskip}
\end{tabular}}

$I_G$(isMarriedTo) = \{isMarriedTo\#1, isMarriedTo\#2\}.

\textbf{Definition}. A generic mapping function $I_{G\_{EXT}}$ is a binary relation that maps a generic property to a set of pairs of resources.

$I_{G\_{EXT}}$: {\it IPg} $\rightarrow$ $2^{IR \times IR}$ such that 

$I_{G\_{EXT}} (p_g) = \{I_{S\_{EXT}} \, (p_s) \, | \, p_s \in $ $I_G(p_g)\}$.

Since {\it IPg} $\subseteq$ {\it IP}, we have 
$I_{G\_{EXT}} (p_g) \subseteq I_{EXT} (p_g)$.


We also have 
$I_{EXT}(p_s)$ = \{$\langle s, o \rangle | \langle s, o \rangle = I_{S\_{EXT}}(p_s)$\}.

$I_{EXT}(p_g)$ = \{$\langle s, o \rangle | \langle s, o \rangle \in I_{G\_{EXT}}(p_g)$\}.

In the example at hand, we have:

\resizebox{0.47\textwidth}{!}{%
\begin{tabular}{ll}  \noalign{\smallskip}
$I_{EXT}$(isMarriedTo\#1) = \{$\langle$BarackObama, MichelleObama$\rangle$\},\\
$I_{EXT}$(isMarriedTo) = \{$\langle$BarackObama, MichelleObama$\rangle$\}. \\ \noalign{\smallskip}
\end{tabular}}

Next, we describe model-theoretic semantics using these mapping functions.


For RDF and RDFS, the entailments are determined by the model-theoretic semantics. For an entailment relation X (e.g., RDFS entailment), an RDF graph $G$ is said to X-entail an RDF graph $G'$ if each X-interpretation that satisfies $G$ also satisfies $G'$. This definition is completed with a mathematical definition of X-interpretation. 

We specify three interpretations: simple, RDF and RDFS by extending the model-theoretic semantics described in \cite{hitzler2011foundations, Nguyen:2014:DLR:2566486.2567973}. For each interpretation, we add additional criteria for supporting the singleton property and the generic property. While we explain the new vocabulary elements in detail, elements without further explanation remain as they are in the original model-theoretic semantics described in \cite{hitzler2011foundations, Nguyen:2014:DLR:2566486.2567973}.

\subsection{Simple Interpretation}
\label{simple}

In the simple interpretation, we formalize the three types of properties as described in Section \ref{concepts}. We also use the mapping functions described in Section \ref{contextual_mapping} to assign each property to a semantic construct.

Given a vocabulary V, the simple interpretation $\mathcal{I}$ consists of:
\begin{enumerate}
\item {\it IR}, a non-empty set of {\it resources}, alternatively called domain or universe of discourse of $\mathcal{I}$,

\item {\it IP}, the set of {\it properties} of  $\mathcal{I}$,
\item {\it IPs}, called the set of {\it singleton properties} of $\mathcal{I}$, as a subset of {\it IP}, 
\item {\it IPg}, called the set of {\it generic properties} of $\mathcal{I}$, as a subset of {\it IP}, {\it IPs} $\cap$ {\it IPg} = $\emptyset$,
\item {\it IPr}, called the set of {\it regular properties},\\ {\it IPr} = {\it IP} $\setminus$ ({\it IPs} $\cup$ {\it IPg}),
\item $I_G$, a function assigning a generic property to a set of its singleton properties,
\item $I_{EXT}$, a mapping function assigning to each property a set of pairs from {\it IR},

 $I_{EXT}$ : {\it IP} $\rightarrow {\it 2^{IR \times IR}}$ where $I_{EXT}(p)$ is called the {\it extension} of property {\it p},
\item $I_{S\_EXT}(p_s)$, the singleton mapping function assigning a singleton property to a pair of resources.

$I_{S\_EXT}$ : {\it IPs} $\rightarrow {\it IR \times IR}$.

\item $I_{G\_EXT}(p_g)$, the generic mapping function assigning each generic property a set of pairs of resources.

$I_{G\_EXT}$ : {\it IPg} $\rightarrow {\it 2^{IR \times IR}}$.

\end{enumerate}


Note that the mapping function $I_{S\_EXT}$ is not a one-to-one mapping; multiple singleton properties may be mapped to the same pair of entities.

\subsection{RDF Interpretation}
\label{rdf}


In the RDF interpretation, we formalize the definition of singleton property, generic property, and how to derive the generic triple from a singleton graph pattern.

{\bf RDF interpretation} of a vocabulary V is a simple interpretation $\mathcal{I}$ of the vocabulary V $\cup$ $V_{RDF}$ that satisfies the criteria from the current RDF interpretation \cite{hayesrdf} and the following criteria:
\begin{enumerate}
\item Define a singleton property\\
{\it $x_s$} $\in$ {\it IPs} iff \\$\langle${\it $x_s$}, rdf:SingletonProperty $^\mathcal{I} \rangle \in I_{EXT} ($rdf:type $^\mathcal{I} \rangle$.


\item Singleton condition\\
If {\it $x_s$} $\in$ {\it IPs} then $\exists! \langle u, v \rangle:  \langle u, v \rangle = I_{S\_EXT}({\it x_s}$), and {\it u,v} $\in$ {\it IR}. This enforces the singleton-ness for the property instances.

\item Define a generic property \\
{\it $x_g$} $\in$ {\it IPg} iff \\$\langle${\it $x_g$}, rdf:GenericProperty $^\mathcal{I} \rangle \in I_{EXT}$ (rdf:type $^\mathcal{I})$. \\
If $x$ $\notin$ {\it IPs}, then {\it IPg} = {\it IPg} $\cup$ \{$x$\}. Any property that is not defined as a singleton property can become generic property. 

\item Infer singleton property and generic property (rule rdf-sp-1 and rdf-sp-2)

{\it $x_s$} $\in$ {\it IPs} and {\it $x_g$} $\in$ {\it IPg} if \\$\langle${\it $x_s$, $x_g$}$\rangle$ $\in I_{\it EXT}($rdf:singletonPropertyOf $^\mathcal{I}$).

A singleton property $x_s$ is connected to a generic property $x_g$ via the property rdf:singletonPropertyOf. As $I_G(x_g)$ is the set of singleton properties connected to the property $x_g$,

$I_G (x_g)= \{x_s\, | \\\, \langle x_s, x_g\rangle \in I_{EXT}($rdf:singletonPropertyOf$^\mathcal{I})\}$.

\item Generic mapping extension

 If $\langle${\it $x_s$}, $x_g \rangle \in I_{EXT}($rdf:singletonPropertyOf $^\mathcal{I}$), then {\it $x_s$} $\in$ {\it IPs}, {\it $x_g$} $\in$ {\it IPg}, and
 ${\mathord I_{S\_EXT}(x_s)} \in I_{G\_EXT}(x_g)$.
$I_{G\_EXT}(x_g)$ is called a {\it generic mapping extension} of the generic property $x_g$. 

$I_{G\_{EXT}} (x_g)= \{I_{S\_{EXT}} \, (x_s) \, | \, x_s \in $ $I_G(x_g)\}$, and

$I_{G\_{EXT}} (x_g) \subseteq I_{EXT} (x_g)$.

\item Generic triple derivation (rule rdf-sp-3)

 If $\langle u, v \rangle = I_{S\_EXT}({x_s})$, and \\$\langle${\it $x_s$}, $x_g \rangle \in I_{EXT}($rdf:singletonPropertyOf $^\mathcal{I}$), \\
then $\langle u, v \rangle \in I_{G\_{EXT}}({x_g})$. 

\textbf{Proof}. $\langle${\it $x_s$}, $x_g \rangle \in I_{EXT}($rdf:singletonPropertyOf $^\mathcal{I}$) implies 
(1): $I_{S\_EXT}({x_s}) \in I_{G\_{EXT}}({x_g})$. 

The combination of $\langle u, v \rangle = I_{S\_EXT}({x_s})$ and (1) implies $\langle u, v \rangle \in I_{G\_{EXT}}({x_g})$. 

This shows how the generic triple $\langle u, v \rangle \in I_{G\_{EXT}}({x_g})$ can be derived from its singleton graph pattern.


\end{enumerate} 


\subsection{RDFS Interpretation}
\label{rdfs}

Here we formalize the connections from a singleton property to its generic property, as well as to other properties such as rdfs:domain, rdfs:range, and rdfs:subPropertyOf. 
We will reuse the function (from \cite{hayesrdf})\\$I_{\it CEXT}: {\it IR} \rightarrow 2^{\it IR}$ 
where $I_{\it CEXT}(y)$ is called a class extension of y, 
$I_{\it CEXT}(y) = \{x \mid \forall x \in {\it IR}: \langle x, y \rangle \in I_{EXT} ($rdf:type$ ^\mathcal{I})\}$.

{\bf RDFS interpretation} of a vocabulary V is an RDF interpretation $\mathcal{I}$ of the vocabulary V $\cup$ $V_{RDFS}$ that satisfies criteria from the current RDFS interpretation \cite{hayesrdf} and the following criteria:

\begin{enumerate}

\item Class rdf:SingletonProperty

\resizebox{0.45\textwidth}{!}{%
\begin{tabular}{ll}
$\langle $rdf:SingletonProperty $^\mathcal{I}, $ rdfs:Class $^\mathcal{I} \rangle \in I_{EXT} ($rdf:type $^\mathcal{I}$). \\ 
\end{tabular}}

The extension of rdf:SingletonProperty class is the set ${\it IPs}$ of all singleton properties, or \\
${\it IPs} = I_{\it CEXT}($rdf:SingletonProperty $^\mathcal{I})$.

\item Class rdf:GenericProperty

\resizebox{0.45\textwidth}{!}{%
\begin{tabular}{ll}
$\langle $rdf:GenericProperty$^\mathcal{I}, $rdfs:Class$^\mathcal{I} \rangle \in I_{EXT} ($rdf:type $^\mathcal{I}$).\\ 
\end{tabular}}

The extension of the rdf:GenericProperty class is the set ${\it IPg}$ of all generic properties, or \\
${\it IPg} = I_{\it CEXT}($rdf:GenericProperty $^\mathcal{I})$.

\item Every singleton property is a resource

$\langle $rdf:SingletonProperty $^\mathcal{I}, $ rdfs:Resource $^\mathcal{I} \rangle \in \\ I_{EXT} ($rdfs:subClassOf $^\mathcal{I}$), 
this causes {\it IPs} $\subseteq$ {\it IR}.

\item Every generic property is a resource

$\langle $rdf:GenericProperty $^\mathcal{I}, $ rdfs:Resource $^\mathcal{I} \rangle \\\in I_{EXT}$$ ($rdfs:subClassOf $^\mathcal{I}$),
this causes {\it IPg} $\subseteq$ {\it IR}.

\item Domain of singleton property (rule rdfs-sp-1)

$\langle x_s, x \rangle \in I_{EXT}($rdf:singletonPropertyOf $^\mathcal{I})$, $\langle x, y \rangle \in I_{EXT}($rdfs:domain $^\mathcal{I})$, 
if $\langle u, v \rangle \in I_{S\_EXT}({x_s})$, 
then $u \in I_{CEXT}(y)$. A singleton property shares the domain with its generic property.

\item Range of singleton property (rule rdfs-sp-2)

$\langle x_s, x \rangle \in I_{EXT}($rdf:singletonPropertyOf $^\mathcal{I})$, $\langle x, y \rangle \in I_{EXT}($rdfs:range $^\mathcal{I})$,
if $\langle u, v \rangle = I_{S\_EXT}({x_s})$, 
then $v \in I_{CEXT}(y)$. A singleton property also shares the range with its generic property.

\item Sub-property condition

If $x,y \in$ {\it IPg}, $\langle${\it $x$}, $y \rangle \in I_{EXT}($rdfs:subPropertyOf $^\mathcal{I}$), then
$I_G (x) \subseteq I_G (y)$, and $I_{G\_{EXT}} (x) \subseteq I_{G\_{EXT}} (y)$.

\item Sub-property upper bound condition


If $y \in$ {\it IPs} and $\langle${\it $x$}, $y \rangle \in I_{EXT}($rdfs:subPropertyOf $^\mathcal{I}$), then $x \in$ {\it IPs} and
$I_{S\_{EXT}} (x) = I_{S\_{EXT}} (y)$. 

\textbf{Proof}. Since $y \in$ {\it IPs}, $\exists! \langle u, v \rangle:  \langle u, v \rangle = I_{S\_EXT}(y)$, and $I_{EXT}(y) = \{\langle u, v \rangle\}$. 

Since $\langle${\it $x$}, $y \rangle \in I_{EXT}($rdfs:subPropertyOf $^\mathcal{I}$), by definition, $I_{EXT}(x) \subseteq I_{EXT}(y) = \{\langle u, v \rangle\}$. 

Since $x$ can be mapped to at most one pair of resources, $x \in$ {\it IPs}, and $I_{S\_{EXT}}(x) = \langle u, v \rangle = I_{S\_{EXT}}(y)$.

\item Sub-property lower bound condition (rule rdfs-sp-4)

If $x \in$ {\it IPg} and $\langle${\it $x$}, $y \rangle \in I_{EXT}($rdfs:subPropertyOf $^\mathcal{I}$), \\then $y \in$ {\it IPg}.

\textbf{Proof}. Assume that $y \in $ {\it IPs}, then $x \in$ {\it IPs} (upper bound condition). We have both $x \in$ {\it IPs} and $x \in$ {\it IPg}. This contradicts to the condition {\it IPs} $\cap$ {\it IPg} = $\emptyset$.

Therefore, $y \notin IPs$, and according to the condition (3) of RDF interpretation, {\it IPg} = {\it IPg} $\cup$ \{$y$\}, or $y \in$ {\it IPg}.

\item Property hierarchy (rule rdfs-sp-3)

 If $\langle x_s$, $x \rangle \in I_{EXT}($rdf:singletonPropertyOf $^\mathcal{I}$), and \\
 $\langle x$, $y \rangle \in I_{EXT}($rdfs:subPropertyOf $^\mathcal{I}$), then 
 $\langle x_s$, $y \rangle \in I_{EXT}($rdf:singletonPropertyOf $^\mathcal{I}$). 
 
 \textbf{Proof}. $\langle x_s$, $x \rangle \in I_{EXT}($rdf:singletonPropertyOf $^\mathcal{I}$) implies (1): $x_s \in I_G (x)$.
 
$\langle x$, $y \rangle \in I_{EXT}($rdfs:subPropertyOf $^\mathcal{I}$) implies \\(2): $I_G (x) \subseteq I_G (y)$.

(1) and (2) derive (3): $x_s \in I_G (y)$. 

In other words, $x_s$ is a singleton property of $y$, or $\langle x_s$, $y \rangle \in I_{EXT}($rdf:singletonPropertyOf $^\mathcal{I}$).


\end{enumerate}


\subsection{OWL 2 RDF-based Semantic Conditions}
\label{owlrules}



 
From the RDF-based semantics of OWL 2 Full \cite{schneider2012owl}, we consider the semantic conditions of the OWL classes and properties that are relevant to singleton properties. These semantic conditions belong to the two categories: logical characteristics of the properties, and relations to other properties. We tighten the semantic conditions of these OWL classes and properties to make sure they are valid in the extended semantics, by enforcing more constraints on the generic property and singleton property extensions. The semantic conditions of these properties must be satisfied in the interpretations extended with singleton property semantics. 

Let V$_p$ be the vocabulary of OWL classes and properties relevant to singleton properties: V$_p$ = \{FunctionalProperty, InverseFunctionalProperty, ReflexiveProperty, IrreflexiveProperty, SymmetricProperty, AsymmetricProperty, TransitiveProperty, inverseOf, equivalentOf\}. 

Let $p_s$, $p'_s$, and $p''_s$ be the singleton properties of the generic property $p$, then $p_s \in I_G(p)$, $p'_s \in I_G(p)$, $p''_s \in I_G(p)$.
 We define the OWL 2 RDF-based interpretation as follows. 
 
{\bf OWL 2 RDF-based interpretation} of a vocabulary V is an RDFS interpretation $\mathcal{I}$ of the vocabulary V $\cup$ $V_{OWL}$ that satisfies criteria from the OWL interpretation \cite{schneider2012owl} and the following semantic conditions: 
\begin{itemize}

\item 
\textbf{Functional property. }If a property is functional, then at most one distinct value can be assigned to any given individual via this property.

A property $p$ is an instance of owl:FunctionalProperty iff $\forall x, y_1, y_2$: 

(1) $p \in IP, \langle x, y_1 \rangle \in I_{EXT}(p), \langle x, y_2 \rangle  \in I_{EXT}(p)$ implies $y_1 = y_2$, 

(2) $p \in IPg, \langle x, y_1 \rangle \in I_{G\_{EXT}}(p), \langle x, y_2 \rangle  \in I_{G\_{EXT}}(p)$ implies $y_1 = y_2$, 

(3) $\forall p_s \in I_G(p), \langle x, y_1 \rangle  = I_{S\_{EXT}}(p_s), \langle x, y_2 \rangle  = I_{S\_{EXT}}(p_s)$ implies $y_1 = y_2$.


\item 
\textbf{Inverse functional property. } An inverse functional property can be regarded as a ``key'' property, i.e., no two different individuals can be assigned the same value via this property. 

A property $p$ is an instance of owl:InverseFunctionalProperty iff $\forall x_1, x_2, y$: 

(1) $p \in IP, \langle x_1, y \rangle \in I_{EXT}(p), \langle x_2, y \rangle  \in I_{EXT}(p)$ implies $x_1 = x_2$, 

(2) $p \in IPg, \langle x_1, y \rangle \in I_{G\_{EXT}}(p), \langle x_2, y \rangle  \in I_{G\_{EXT}}(p)$ implies $x_1 = x_2$, 

(3) $\forall p_s \in I_G(p), \langle x_1, y \rangle  = I_{S\_{EXT}}(p_s), \langle x_2, y  \rangle  = I_{S\_{EXT}}(p_s)$ implies $x_1 = x_2$.


\item
\textbf{Reflexive property. } A reflexive property relates every individual in the universe to itself.
 
A property $p$ is an instance of the class owl:ReflexiveProperty iff  $\forall x$: 

(1) $p \in IP , \langle x , x \rangle \in I_{EXT}(p)$, 

(2) $p \in IPg , \langle x , x \rangle \in I_{G\_{EXT}}(p)$, 

(3) $\forall p_s \in I_G(p) , \langle x , x \rangle = I_{S\_{EXT}}(p_s)$.


\item
\textbf{Irreflexive property. } An irreflexive property does not relate any individual to itself.

A property $p$ is an instance of the class owl:IrreflexiveProperty iff  $\forall x$: 

(1) $p \in IP , \langle x , x \rangle \notin I_{EXT}(p)$, 

(2) $p \in IPg , \langle x , x \rangle \notin I_{G\_{EXT}}(p)$, 

(3) $\forall p_s \in I_G(p) , \langle x , x \rangle \neq I_{S\_{EXT}}(p_s)$.


\item
\textbf{Symmetric property. } If two individuals are related by a symmetric property, then this property also relates them reversely.

A property $p$ is an instance of the class owl:SymmetricProperty iff $\forall x , y :$ 

(1) $p \in IP,  \langle x, y \rangle \in I_{EXT}(p)$ implies $ \langle y, x \rangle \in I_{EXT}(p)$,

(2) $p \in IPg,  \langle x, y \rangle \in I_{G\_{EXT}}(p)$ implies $ \langle y, x \rangle \in I_{G\_{EXT}}(p)$,

(3) $\forall p_s \in I_G(p),  \langle x, y \rangle = I_{S\_{EXT}}(p_s) $ implies $ \exists p'_s \in I_G(p), \langle y, x \rangle = I_{S\_{EXT}}(p'_s)$.


\item
\textbf{Asymmetric property. } If two individuals are related by an asymmetric property, then this property never relates them reversely.

A property $p$ is an instance of the class owl:AsymmetricProperty iff $\forall x, y:$

(1) $p \in IP,  \langle x, y \rangle \in I_{EXT}(p)$ implies $ \langle y, x \rangle \notin I_{EXT}(p)$,

(2) $p \in IPg,  \langle x, y \rangle \in I_{G\_{EXT}}(p)$ implies $ \langle y, x \rangle \notin I_{G\_{EXT}}(p)$,

(3) $\forall p_s \in I_G(p),  \langle x, y \rangle = I_{S\_{EXT}}(p_s) $ implies $\nexists p'_s \in I_G(p),  \langle y, x \rangle = I_{S\_{EXT}}(p_s)$.


\item 
\textbf{Transitive property.} A transitive property that relates an individual $a$ to an individual $b$, and individual $b$ to an individual $c$, also relates $a$ to $c$.

A property $p$ is an instance of the class owl:TransitiveProperty iff $\forall x, y, z: $

(1) $p \in IP, \langle x, y \rangle \in I_{EXT}(p), \langle y, z \rangle \in I_{EXT}(p)$ implies $ \langle x, z \rangle \in I_{EXT}(p)$,

(2) $p \in IPg, \langle x, y \rangle \in I_{G\_{EXT}}(p), \langle y, z \rangle \in I_{G\_{EXT}}(p)$ implies $ \langle x, z \rangle \in I_{G\_{EXT}}(p)$,

(3) $\forall p_s, p'_s \in I_G(p), \langle x, y \rangle = I_{S\_{EXT}}(p_s), \langle y, z \rangle = I_{S\_{EXT}}(p'_s)$ implies $\exists p''_s \in I_G(p),\langle x, z \rangle = I_{S\_{EXT}}(p''_s)$.


\item
\textbf{Inverse property}. The inverse of a given property is the corresponding property with subject and object swapped for each property assertion built from it.

$( p_1 , p_2 ) \in I_{EXT}($owl:inverseOf $^\mathcal{I}$) iff 

(1) $ \, \forall p_1, p_2 \in IP, I_{EXT}(p_1) = \{ \langle x, y \rangle \, | \, \langle y, x \rangle \in I_{EXT}(p_2) \}$,

(2) $ \, \forall p_1, p_2 \in IPg, I_{G\_{EXT}}(p_1) = \{ \langle x, y \rangle \, | \, \langle y, x \rangle \in I_{G\_{EXT}}(p_2) \}$,

(3) $ \, \forall p_s \in I_G(p_1), \forall p'_s \in I_G(p_2), I_{S\_{EXT}}(p_s) = \langle x, y \rangle \,$, $I_{S\_{EXT}}(p'_s) = \langle y, x \rangle$.


\item
\textbf{Equivalent property}. Two equivalent properties share the same property extension.

$( p_1 , p_2 ) \in I_{EXT}($owl:equivalentOf $^\mathcal{I}) $ iff 

(1) $ \forall p_1, p_2 \in IP: I_{EXT}(p_1) = I_{EXT}(p_2)$,

(2) $ \forall p_1, p_2 \in IPg: I_{G\_{EXT}}(p_1) = I_{G\_{EXT}}(p_2)$,

(3) $ \, \forall p_s \in I_G(p_1), \exists p'_s \in I_G(p_2): I_{S\_{EXT}}(p_s) = I_{S\_{EXT}}(p'_s)$.

\end{itemize}


\section{Contextual Inferences}
\label{rules}

In the RDF, RDFS, and OWL 2 Full interpretations, we have proved several deduction rules in Section \ref{formal}.
Here we present a set of these rules in Section \ref{entailment-rules} and demonstrate how these rules can be applied to derive the inferred statements described in the motivating example in Section \ref{reasoning-types}.

\subsection{Contextual Entailment Rules}
\label{entailment-rules}
The three following rdf-sp rules are derived from the RDF interpretation.
\[
\inference {\predicate{$u$  rdf:singletonPropertyOf $v$}\quad .  }{\predicate{$u$ rdf:type rdf:SingletonProperty \quad . } }[(rdf-sp-1)]
\]
\[
\inference {\predicate{$u$  rdf:singletonPropertyOf $v$} \quad . }{\predicate{$v$ rdf:type rdf:GenericProperty \quad . } }[(rdf-sp-2)]
\]
\[
\inference {\predicate{$u$  rdf:singletonPropertyOf $v$} \quad . \\ \quad \predicate{$x$ $u$ $y$}\quad . }{ \predicate{ $x$ $v$ $y$ \quad . }}[(rdf-sp-3)]
\]
The four rdfs-sp rules are derived from the RDFS interpretation.
\[
\inference {\predicate{$u$  rdf:singletonPropertyOf $v$} \quad . \\ \quad \predicate{$v$ rdfs:domain $x$}\quad . }{\predicate{$u$ rdfs:domain $x$}\quad . }[(rdfs-sp-1)]
\]
\[
\inference {\predicate{$u$  rdf:singletonPropertyOf $v$} \quad . \\ \quad \predicate{$v$ rdfs:range $y$}\quad . }{\predicate{$u$ rdfs:range $y$}\quad . }[(rdfs-sp-2)]
\]
\[
\inference { \predicate{$u$  rdf:singletonPropertyOf $x$} \quad . \\ \predicate{$x$  rdfs:subPropertyOf $y$} \quad. }{ \predicate{$u$  rdf:singletonPropertyOf $y$} \quad . \quad}[(rdfs-sp-3)]
\]
\[
\inference { \predicate{$x$  rdf:type rdf:GenericProperty} \quad . \\ \predicate{$x$  rdfs:subPropertyOf $y$} \quad. }{ \predicate{$y$  rdf:type GenericProperty} \quad . \quad}[(rdfs-sp-4)]
\]
\[
\inference { \predicate{$u$  rdf:singletonPropertyOf $x$} \quad . \\ \predicate{$x$  rdfs:subPropertyOf $y$} \quad. }{ \predicate{$y$  rdf:type GenericProperty} \quad . \quad}[(rdfs-sp-5)]
\]

The rule rdfs-sp-5 can easily be drived by combining the two rules rdfs-sp-3 and rdf-sp-2. Similar to the property rdfs:subPropertyOf, here we also provide the rules for the owl:equivalentOf.
\[
\inference { \predicate{$u$  rdf:singletonPropertyOf $x$} \quad . \\ \predicate{$x$  owl:equivalentOf $y$} \quad. }{ \predicate{$u$  rdf:singletonPropertyOf $y$} \quad . \quad}[(owl-sp-1)]
\]
\[
\inference { \predicate{$x$  rdf:type GenericProperty} \quad . \\ \predicate{$x$  owl:equivalentOf $y$} \quad. }{ \predicate{$y$  rdf:type GenericProperty} \quad . \quad}[(owl-sp-2)]
\]
\[
\inference { \predicate{$u$  rdf:singletonPropertyOf $x$} \quad . \\ \predicate{$x$  owl:equivalentOf $y$} \quad. }{ \predicate{$y$  rdf:type GenericProperty} \quad . \quad}[(owl-sp-3)]
\]

If $x$ owl:equivalentOf $y$ then \\(1) $x$ rdfs:subPropertyOf $y$ and \\(2) $y$ rdfs:subPropertyOf $x$.

The above three owl-sp rules can be derived easily by combing this rule and the rules rdfs-sp-3, rdfs-sp-4, and rdfs-sp-5, respectively.



\subsection{Contextual Inferencing}
\label{reasoning-types}

Back to the motivating example, from the contextual statement ``BarackObama is married to Michelle Obama in Chicago'', we as humans can infer a list of statements ($S_1$ to $S_5$) as shown in Table \ref{tab:derived_statements}. Here we demonstrate step-by-step how to infer statements $S_1$ to $S_5$ from the original statement. 

Our initial knowledge base includes the singleton property graph pattern representing the original contextual statement and the background knowledge as follows:

\resizebox{0.47\textwidth}{!}{%
\begin{tabular}{l@{\hspace{1em}}l @{\hspace{1em}}l @{\hspace{1em}}l}  \noalign{\smallskip}
$SP_1$: & BarackObama & isMarriedTo\#1 & MichelleObama .\\
$SP_2$: & isMarriedTo\#1 & singletonPropertyOf & isMarriedTo .\\
$SP_3$: & isMarriedTo\#1 & happenedIn & Chicago .\\ \noalign{\smallskip}
$T_2$: & isMarriedTo & subPropertyOf & isSpouseOf .\\
$T_5$: & Chicago & partOf & Illinois .\\
$T_6$: & Illinois & partOf & USA . \\\noalign{\smallskip}
\end{tabular}}

Assume that we also have a partOf-rule that, if $x$ happened in a place $y$ which is a part of a bigger place $z$, then $x$ also happened at the place $z$.
\[
\inference { \predicate{$x$  happenedIn $y$} \quad . \quad\predicate{$y$  partOf $z$} \quad. }{ \predicate{$x$  happenedIn z} \quad . \quad}[(partOf-rule)]
\]

We start by applying the partOf-rule on the triples $SP_3$ and $T_5$, we obtain the triple $SP_4$.

\resizebox{0.40\textwidth}{!}{%
\begin{tabular}{l@{\hspace{1em}}l @{\hspace{1em}}l @{\hspace{1em}}l}  \noalign{\smallskip}
$SP_4$: & isMarriedTo\#1 & happenedIn & Illinois .\\ \noalign{\smallskip}
\end{tabular}}

The singleton triple pattern including triples $SP_1$ and $SP_2$ derives the statement ``Barack Obama is married to Michelle Obama'' according to rule rdf-sp-3. Combining three triples $SP_1$, $SP_2$, and $SP_4$ will derive the contextual statement $S_1$ ``Barack Obama isMarriedTo Michelle Obama in Illinois''.

Re-applying the partOf-rule on the triples $SP_4$ and $T_6$, we obtain the new triple $SP_5$.

\resizebox{0.40\textwidth}{!}{%
\begin{tabular}{l@{\hspace{1em}}l @{\hspace{1em}}l @{\hspace{1em}}l}  \noalign{\smallskip}
$SP_5$: & isMarriedTo\#1 & happenedIn & USA .\\ \noalign{\smallskip}
\end{tabular}}

Similar to $S_1$, the combination of triples $SP_1$, $SP_2$, and $SP_5$ derives the contextual statement $S_2$ ``Barack Obama isMarriedTo Michelle Obama in USA''.

Next, if we apply the rule rdfs-sp-3 on the triples $SP_2$ and $T_2$, we obtain the new triple $SP_6$.

\resizebox{0.47\textwidth}{!}{%
\begin{tabular}{l@{\hspace{1em}}l @{\hspace{1em}}l @{\hspace{1em}}l}  \noalign{\smallskip}
$SP_6$: & isMarriedTo\#1 & singletonPropertyOf & isSpouseOf .\\ \noalign{\smallskip}
\end{tabular}}

Applying the rule rdf-sp-3 on the triples $SP_1$ and $SP_6$ derives the statement ``Barack Obama isSpouseOf Michelle Obama''.

The combination of the triples $SP_1$, $SP_6$, and $SP_3$ will derive the contextual statement $S_3$ ``Barack Obama isSpouseOf Michelle Obama in Chicago''.

Similarly, combining the triples $SP_1$, $SP_6$, and $SP_4$ will derive the contextual statement $S_4$ ``Barack Obama isSpouseOf Michelle Obama in Illinois''. 

The combination of $SP_1$, $SP_6$, and $SP_5$ derives the contextual statement $S_5$ ``Barack Obama isSpouseOf Michelle Obama in USA''.

Therefore, we have shown how the contextual statements $S_1$ to $S_5$ can be inferred in our approach.

\section{Implementation}
\label{implementation}
Here we explain how we compute all inferred triples using the proposed entailment rules for existing knowledge bases in two steps. First we describe how we transform the existing knowledge bases into the singleton property representation in Section \ref{transform}. Then Section \ref{compute} describes how all the proposed rules are computed for every triple in the resulting knowledge bases.

\subsection{Transforming Representation}
\label{transform}

As we discussed earlier in Section \ref{introduction}, knowledge bases like  DBpedia and Bio2RDF represent the contextual information such as provenance in the form of a quad. Before computing the inferred triples for these datasets, we need to prepare the knowledge bases by transforming them to the singleton property representation.

Given any quad in the form of ($s$, $p$, $o$, $g$), we transform it to the singleton property representation by creating a singleton property ($sp_i$, singletonPropertyOf, $p$) and asserting the singleton triple ($s$, $sp_i$, $o$). We use the property \verb|wasDerivedFrom| from the PROV ontology \cite{lebo2012prov} to represent the provenance of the triple ($sp_i$, prov:wasDerivedFrom, $g$). The singleton property URIs are constructed by appending a unique string to the generic property URI, with an incremental counter for the unique number in the whole dataset.

We developed a Java 8 tool, called rdf-contextualizer, to transform any RDF datasets from the quad representation to the singleton property. We took advantages of the Jena RIOT API \cite{jena-riot} with high throughput parsers for parsing an input file from any RDF format and generating a stream of RDF quads. For each quad stream, we created a pipeline of streams for converting each quad to the singleton property representation, shortening triples to Turtle format, and writing them to gzip files though buffer writers. As each stream is handled by a separate thread, we can utilize the CPU resources, especially the ones with multiple cores, by creating multiple threads for parsing multiple files concurrently.

We validated the syntax of the output datasets by writing an analyzer to parse the output files and also generate the statistics reported in Section \ref{results}.

\subsection{Computing Inferred Triples}
\label{compute}

Running all contextual entailment rules on every singleton triple produces at least two more triples (rdf-sp-1 and rdf-sp-3). The number of inferred triples goes up to multi-billion with datasets like  DBpedia and Bio2RDF. That amount of inferred triples cannot fit an in-memory reasoner such as Jena \cite{carroll2004jena}. The proposed entailment rules can also be computed in the reasoners with the support for user-defined rules such as Oracle
\cite{wu2008implementing}. However, for the rules generating a large number of inferred triples, optimization is necessary as Oracle has optimized the computation of large number of inferred triples for owl:sameAs \cite{kolovski2010optimizing}.
Without taking such optimization step in the existing engines, it is time-consuming to query the rule patterns and insert the inferred triples to the store because the insert query is always expensive. 
Therefore, the proposed contextual entailment rules may not be best computed in the existing engines without taking the optimization step. 

To demonstrate that the computation of the proposed rules can be scalable with proper optimization, we implemented our engine in the tool rdf-contextualizer. Since the proposed rules can be applied to each triple independently, we can pass the triples to concurrent tasks. While transforming the RDF quads to the singleton property representation, we added an optional inferring task in the stream pipelines to compute the proposed rules for every triple. The time difference between the runs with and without the \textit{-infer} option would be the run time for inferring triples added to the overall process.

\section{Evaluation}
\label{evaluation}
We evaluate the performance of computing inferred triples from the proposed rules on a large scale by using the tool rdf-contextualizer on real-world knowledge bases.
\subsection{Experiment Setup}

We use a single server installed with Ubuntu 12.04. It has 24 cores, each core is Intel Xeon CPU 2.60GHz. We use two disks, one SSD 220GB for storing input datasets, and one HD disk 2.7T for writing the output. This server has 256GB of RAM, however, we limit 60GB for each Java program.

\subsection{Datasets}

We downloaded and used the ontologies and RDF quad datasets from  DBpedia \cite{dbpedia} and four Bio2RDF datasets including NCBI Genes \cite{ncbigene}, PharmGKB \cite{pharmgkb}, CTD \cite{ctd}, and GO Annotations \cite{goa} in our evaluation. We chose these quad datasets because they are large and widely-used with high impact in the community. For Bio2RDF datasets, we also downloaded the Bio2RDF mapping files \cite{bio2rdf-mapping}. 
\begin{table}[h!t]
\centering
\caption{Number of RDF quads in the original datasets and number of unique RDF quads in the duplicate-removed datasets}
\label{datasets}
\begin{tabular}{|l|r|r|}\noalign{\smallskip}\hline 
Dataset & \# of Quads & \# of Unique Quads \\ \hline
NCB-NG     & 4,043,516,408     & 2,010,283,374                        \\ \hline
DBP-NG     & 1,039,275,891     & 784,508,538                          \\ \hline
CTD-NG     & 644,147,853       & 327,648,659                          \\ \hline
PHA-NG     & 462,682,871       & 339,058,720                          \\ \hline
GOA-NG     & 159,255,577       & 97,522,988                         \\ \hline \noalign{\smallskip}
\end{tabular}
\end{table}

We reported the number of RDF quads per dataset in the first and second columns of Table \ref{datasets}. The dataset identifier is taken from the first 3 letters of its name. 

We observed that there were too many duplicate quads among the files within each dataset. We also believe that the duplicates may be created on purpose. Each of these datasets has a number of files and each file contains a number of RDF quads for a topic. For example, NCBI Genes dataset has one file for all genes belonging to one species. Therefore, we keep these datasets in the original version and created a new version for each dataset with all duplicates being removed. We removed the duplicates by 1) concatenating all files into a single file, 2) splitting this file into multiple smaller files, 3) sorting each small file, and 4) merging all the sorted files into a single file. The third column of Table \ref{datasets} shows the number of unique quads per dataset after removing the duplicates.

We then generated the singleton property version of each dataset by running the tool rdf-contextualizer with and without the \textit{-infer} option. We ran each dataset version at least 3 times and reported the average results in Section \ref{results}. 

The tool rdf-contextualizer and the materials used in this paper are publicly available for reproducing the experiments\footnote{\url{https://archive.org/services/purl/rdf-contextualizer}}.

\subsection{Results}
\label{results}
\begin{figure*}[h!t]
  \centering
      \includegraphics[width=0.8\textwidth]{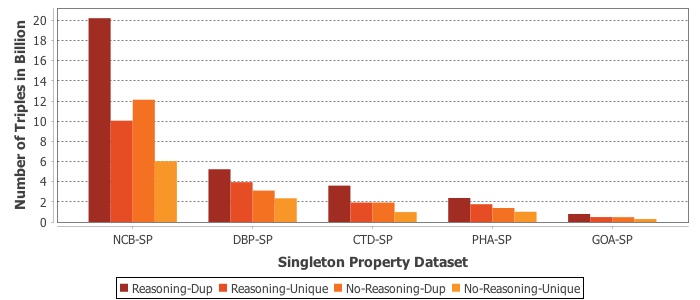}
  \caption{Total number of triples for each dataset in all four cases: with vs. without reasoning and with vs. without removing duplicates. \label{data-chart}}
\end{figure*}

\begin{figure*}[h!t]
  \centering
      \includegraphics[width=0.8\textwidth]{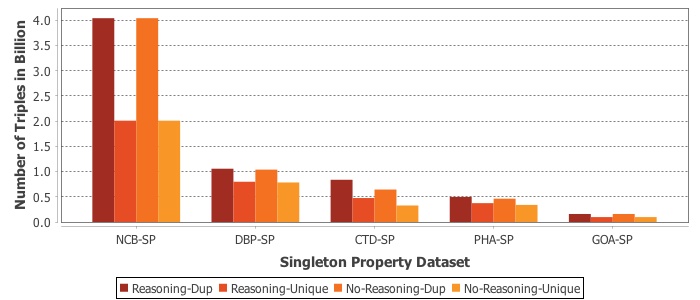}
  \caption{Total number of singleton triples for each dataset in all four cases: with vs. without reasoning and with vs. without removing duplicates. \label{singleton-chart}}
\end{figure*}

We consider four dimensions in our evaluation: number of triples, number of singleton triples, run time, and disk space.
In all figures, the series \verb|Reasoning| stands for running the tool with \textit{-infer} option and \verb|No-Reasoning| stands for running the tool without \textit{-infer} option. Running the \verb|NoReasoning| option provides the results as the baseline cost. The differences in the results between the \verb|NoReasoning| and \verb|Reasoning| versions are the extra cost estimated for the computation of the inferred triples. We run the tool on two versions of datasets. The series with \verb|Dup| denotes the datasets with duplicate quads and the series with \verb|Unique| denotes the datasets with all duplicates being removed. 

Combining the two options: with vs. without \textit{-infer} and with vs. without removing duplicates, each evaluating dimension has four cases: \verb|Reasoning-Dup|, \verb|Reasoning-Unique|, \verb|NoReasoning-Dup|, and \verb|NoReasoning-Unique|.

\subsubsection{Number of Inferred Triples} Figure \ref{data-chart} shows the total number of triples of each dataset in four cases.

In general, the \verb|Reasoning| cases contain larger number of triples than the \verb|NoReasoning| cases, from 66.7\% to 96.67\%. For example, the \verb|Reasoning-Dup| version of NCBI genes, the largest dataset, contains about 8 billion inferred triples (66.67\%) more than the \verb|NoReasoning-Dup| version. Similarly, the \verb|Reasoning-Unique| version of NCBI genes contains about 4 billion inferred triples more than the \verb|NoReasoning-Unique| version. Some other datasets have their number of inferred triples higher than 66.67\%. For example, the \verb|Reasoning-Unique| version of CTD contains about 950 million inferred triples (96.67\%) more than the \verb|NoReasoning-Unique| version.

Between the \verb|Dup| and \verb|Unique| versions, the \verb|Reasoning-Dup| versions contain 50\% (NCBI Genes), 32\% ( DBpedia), 86\% (CTD), 35\% (PharmGKB) and 63\% (GOA) more triples than the corresponding \verb|Reasoning-Unique| versions. 

\begin{figure*}[h!t]
  \centering
      \includegraphics[width=0.8\textwidth]{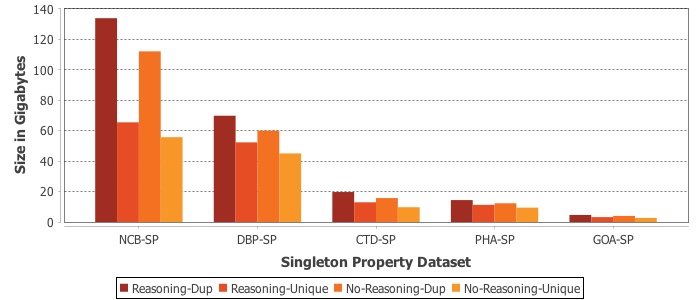}
  \caption{Disk space (zipped) for each dataset in all four cases: with vs. without reasoning and with vs. without removing duplicates. \label{disk-chart}}
\end{figure*}

\begin{figure*}[h!t]
  \centering
      \includegraphics[width=0.8\textwidth]{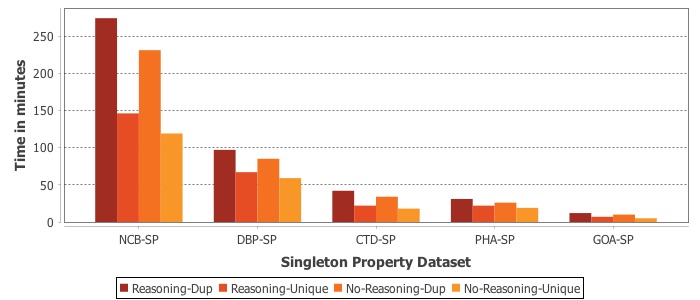}
  \caption{Run time (in minutes) for each dataset in all four cases: with vs. without reasoning and with vs. without removing duplicates. \label{time-chart}}
\end{figure*}

\subsubsection{Number of Inferred Singleton Triples} Resulted from the inference rules involving property hierarchy in the schema, the number of inferred singleton triples varies across datasets as shown in Figure \ref{singleton-chart} since they have different schema. For example, since NCBI Genes dataset does not have property hierarchy, the number of singleton triples remains the same in both version \verb|Reasoning| and \verb|NoReasoning|. Meanwhile, CTD dataset inferred 194 million singleton triples (30\%) for the \verb|Reasoning-Dup| version and 148 million singleton properties (45\%) for the \verb|Reasoning-Unique| version.

\subsubsection{Disk Space} 
Generally speaking, for every RDF quad, we generated 3 triples for \verb|NoReasoning| version and at least 5 triples for the \verb|Reasoning| version. That makes the overall number of triples of each SP dataset increase up to 3-5 times compared to the number of quads of the corresponding named graph dataset. Consequently, the disk size of one SP dataset could be 3-5 times more than the disk size of its corresponding named graph dataset. This is the case in which the triples of SP datasets being serialized into the N-Triple format. For very large datasets like  DBpedia and NCBI Genes, we do not recommend this N-Triple serialization since the unzipped files may require disks with tetrabyte capacity.

RDF Turtle format is more compact than the N-Triple format, especially with very large datasets. Since our approach enables data to be represented in the RDF triple form, we chose the Turtle format to serialize the triples to files. It allows us to arrange the triple order so that we can shorten the strings of triples sharing subject, or subject and predicate. For shortening the URIs in the Turtle format, we compiled a list of prefixes used in these datasets. 
Thanks to this compact representation, comparing to the \verb|NoReasoning-Dup| version, the \verb|Reasoning-Dup| version only adds 15-25\% more space for 66.7-96.97\% more number of triples (Figure \ref{disk-chart}). 

\subsubsection{Run Time} 

Figure \ref{time-chart} shows the run time execution for all four cases of each dataset. GO Annotations dataset, took 11 minutes for 480 million triples of \verb|NoReasoning-Dup| version and 13 minutes for 796 million triples of \verb|Reasoning-Dup| version. In other words, it added 2 minutes to the overall process to infer 316 million triples. NCBI Genes, the largest dataset, took 232 minutes for 12 billion triples of \verb|NoReasoning-Dup| version and 274 minutes for 20 billion triples of \verb|Reasoning-Dup| version. In terms of percentage, the \verb|Reasoning-Dup| versions of these datasets add 14-19\% run time for inferring 66.7-96.67\% of the total number of triples in the \verb|NoReasoning-Dup| versions. This shows that the inferencing time is quite small and practical.

\begin{figure}[h!t]
  \centering
      \includegraphics[width=0.48\textwidth]{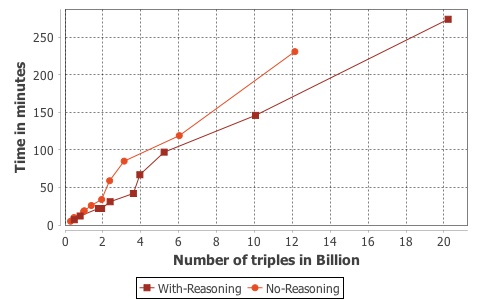}
  \caption{ Run time (in minutes) across datasets in two cases: with vs. without reasoning. \label{time_size-chart}}
\end{figure}

\begin{figure}[h!t]
  \centering
      \includegraphics[width=0.48\textwidth]{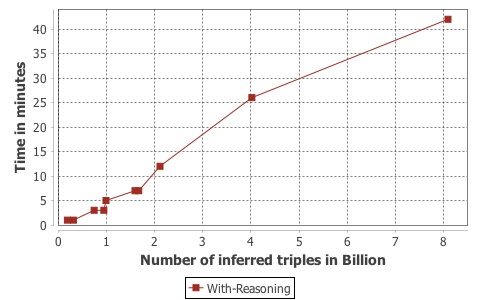}
  \caption{Run time (in minutes) for inferred triples. \label{time_inferred-chart}}
\end{figure}

We plot the size of the datasets and the time execution in the same chart to show the correlation between them. Figure \ref{time_size-chart} shows that when the size of the datasets increases, the time execution for both \verb|Reasoning| and \verb|NoReasoning| case also increases almost linearly to the size of the dataset. This figure also shows that for the same number of triples, the \verb|Reasoning| versions take shorter run time than the \verb|NoReasoning| versions. This is reasonable because for generating the same amount of triples, the time it takes for the \verb|NoReasoning| versions to parse content from files and serialize the output to files is longer than the time it does for the \verb|Reasoning| version to infer the triples. 
Figure \ref{time_inferred-chart} plots the time execution and the number of inferred triples. It also shows the run time is linear to the number of inferred triples.

These results are consistent with our Java stream-based implementation. Since each triple can be processed independently, a set of triples can be passed to multiple streams for concurrent processing instead of sequential processing. Multi-core CPUs are also utilized for processing multiple files concurrently. 

\subsubsection{Overall Remarks} 

Between the \verb|Reasoning| and \verb|NoReasoning| versions, the results show that the numbers of triples of the \verb| Reasoning| versions are higher than the number of triples in the \verb|NoReasoning| versions from 66.7\% to 96.67\%. In terms of number of inferred singleton triples, the \verb|Reasoning| versions produce up to 45\% of the number of singleton triples in the \verb|NoReasoning| version. In terms of disk space, the \verb|Reasoning| versions require 15-25\% of disk space of the \verb|NoReasoning| version. Last but not least, the \verb|Reasoning| versions added 14-19\% run time to the overall process.

Between the \verb|Dup| and \verb|Unique| versions, the results show that the number of triples, number of singleton triples, the disk space, and the run time reduce significantly (up to 50\%) in the \verb|Unique| versions.



\section{Related Work}
\label{related}
Representing and querying contextual information about triples has received significant attention, and several approaches have been proposed. We can classify these approaches into three categories: triple (reification, singleton property), quadruple (named graph), and quintuple (RDF+ \cite{schueler2008querying}). 
However, logical inferences with contextual information about triples remain largely underdeveloped due to the lack of a model-theoretic semantics that would determine entailment rules. Without such a model-theoretic semantics, we can make up some rules using the syntax of RDF reification to simulate our proposed rules. Nevertheless, these syntactical rules are not logically valid since they are neither logically derived from nor proven in a model-theoretic semantics. Therefore, we chose the singleton property approach over other approaches to develop the proposed inferencing mechanism mainly because it comes with a formal semantics. 
To the best of our knowledge, our proposal is the first one to provide the model-theoretic semantics with entailment rules that enables the entailment of new contextual triples about triples.



The stream reasoning \cite{nguyen2013slubm} where the temporal dimension is not represented directly in RDF may benefit from our work as it allows the temporal dimension to be incorporated within the RDF syntax. The temporal RDF \cite{gutierrez2005temporal} incorporates temporal reasoning into RDF using reification. This temporal RDF may take advantages of singleton property semantics as temporal information can be incorporated into RDF through singleton properties instead of reification. 

\section{Discussion and Future Work}
\label{discuss}
\textbf{Applications.} 
We have implemented and evaluated the proposed rules with the forward-chaining inferences in this paper. The backward chaining inferences with the proposed rules can be implemented in the reasoners. They can also be implemented in the triple stores for answering SPARQL queries based on the triples inferred from the proposed rules. We also believe that the proposed inference mechanism will benefit several applications, such as streaming reasoning, temporal reasoning, tracking context of inferred triples, and question answering based on extracted knowledge and logical inferences.

\textbf{Performance}. Performance is the real challenge for Semantic Web reasoning in general, and also for the proposed inferencing mechanism, especially on the Web scale. Inferred triples can be computed by SPARQL INSERT query to find the rule patterns and insert the matching triples to triple stores. However, this approach is not scalable as insert query is costly. Computing the inferred triples on a large scale requires the reasoners to be optimized. Since the singleton pattern is fixed, it can be indexed for faster retrieval. Furthermore, our evaluation shows that parallelizing the computation such as using the stream-based pipelines would also improve the performance for existing reasoners.


\textbf{OWL 2.} We have studied RDF-based semantics of the OWL 2 Full and obtained initial results on its compability with the semantics we proposed here because they are based on the model theory. We believe that the contextual inferences can also be applicable to OWL 2 DL and OWL 2 direct semantics. However, we need to extend the semantics of these OWL 2 profiles with new semantic constructs in order to accomodate the proposed conceptual model.

\textbf{Formal studies.} Incorporating these contextual information into RDF triples would enable several logics such as temporal reasoning, geological reasoning, and provenance reasoning to be studied and applied in these knowledge bases. Logical reasoning tasks such as consistency checking, classification, subsumption, or deriving new knowledge would allow more intelligent systems to be developed, especially on the Web scale.

\section{Conclusion}
\label{conclude}

We have presented our inferencing mechanism that allows contextual statements, in the form of RDF contextual triples about triples, to be reasoned with. Our proposed mechanism is theoretically sound and computationally scalable. Our model-theoretic semantics represents the contextual statements as first-class citizens and enables them to be inferred with the proposed entailment rules. We also demonstrated the feasibility and scalability of computing inferred triples using the proposed entailment rules in various real-world knowledge bases.



\bibliographystyle{abbrv}
\bibliography{all,knowledge}

\end{document}